# 基于 VMD 和 IPSO-ELM 的短期负荷预测模型研究

谢 强

(湖南工业大学电气与信息工程学院，湖南 株洲 412000)

**摘要**：为提高风电场的电力负荷预测精度，本研究提出了一种结合变分模态分解（Variational Mode Decomposition, VMD）与改进的粒子群优化算法（Improved Particle Swarm Optimization, IPSO）来优化极限学习机（Extreme Learning Machine, ELM）的先进组合预测方法。首先，利用 VMD 算法对电力负荷原始数据进行高精度的模态分解，并基于互信息熵理论将分解结果划分为高频和低频序列。然后，通过整合 Tent 混沌映射，本研究对传统 PSO 算法进行了深度改进，提出 IPSO-ELM 预测模型。该模型对高频与低频序列进行独立预测并进行数据重构，以获取最终的预测结果。仿真结果表明，与传统 ELM、MVO-ELM、PSO-ELM 方法相比，所提方法在预测精度和收敛速度上展现出了显著的改进。

**关键词**：负荷预测；变分模态分解；极限学习机；改进粒子群优化；Tent 混沌映射

**中图法分类号**:TM715　　　　**文献标识码**: A

## Research on short-term load forecasting model based on VMD and IPSO-ELM

Qiang Xie

(College of Electrical and Information Engineering, Hunan University of Technology, Zhuzhou 412000, China)

**Abstract**: To enhance the accuracy of power load forecasting in wind farms, this study introduces an advanced combined forecasting method that integrates Variational Mode Decomposition (VMD) with an Improved Particle Swarm Optimization (IPSO) algorithm to optimize the Extreme Learning Machine (ELM). Initially, the VMD algorithm is employed to perform high-precision modal decomposition of the original power load data, which is then categorized into high-frequency and low-frequency sequences based on mutual information entropy theory. Subsequently, this research profoundly modifies the traditional multiverse optimizer by incorporating Tent chaos mapping, exponential travel distance rate, and an elite reverse learning mechanism, developing the IPSO-ELM prediction model. This model independently predicts the high and low-frequency sequences and reconstructs the data to achieve the final forecasting results. Simulation results indicate that the proposed method significantly improves prediction accuracy and convergence speed compared to traditional ELM, PSO-ELM, and PSO-ELM methods.

## 0 引言

由于风力发电的输出波动性和不可预测性可能导致电网并网时的安全风险，因此对风电场的风速、功率和负荷等数据进行精准预测显得尤为重要[1-3]。提升电力负荷预测精度对于电网的优化调度[4]、降低发电成本以及提高系统的可靠性等方面都具有关键作用[5-6]。

在负荷预测领域，通常依赖历史数据和各种预测模型来进行预测，包括随机森林、

神经网络、支持向量机和极限学习机等[7-12]。为了提升这些模型的预测准确性，研究通常集中在模型改进和参数优化两大方向。例如，基于传统神经网络提出了全局时延反馈神经网络、卷积神经网络、模糊神经网络等改进型模型[13-15]，并结合群智能算法如布谷鸟算法、粒子群优化和蚁群算法来优化模型参数[16-20]。粒子群优化算法（Particle Swarm Optimization，PSO）是一种模拟鸟群社会行为的群智能优化算法，以其参数配置简便、鲁棒性好和快速收敛的特性而被广泛应用于多个领域[21-24]。例如，PSO 在优化支持向量机的参数来提高预测性能方面[25]。尽管 PSO 算法在执行速度和实现简易性方面具有优势，它也存在搜索过程中可能陷入局部最优解和全局搜索能力不足的问题[26-28]。此外，原始预测数据的预处理也是一个重要的研究方向，方法如小波变换、经验模态分解以及变分模态分解等[29-30]。变分模态分解在处理数据时能够自适应地匹配每个模态的最佳中心频率和带宽，有效地避免了模态之间的混叠。

为此，本研究首先对 PSO 算法进行改进，引入 Tent 混沌映射进行种群初始化，提升算法性能。接着，采用变分模态分解对原始数据进行预处理，分离出高频和低频序列，简化了数据复杂度。通过这一系列优化，开发了 VMD-IPSO-ELM 组合预测模型，并通过仿真分析与其他算法进行比较，验证了所提模型的高效性和准确性。

# 1 变分模态分解

## 1.1 原理概述

假设 $v_k(t)$ 是原始信号分解后的模态分量，需满足各个 $v_k(t)$ 的带宽估计之和最小，各模态之和与原始信号 $f$ 相等，则具体构造变分问题如下[31]：

$$\begin{cases} \min_{\{v_k\},\{\omega_k\}} \left\{ \sum_k \left\| \partial_t \left[ (\delta(t) + j/\pi t) * v_k(t) \right] e^{-j\omega_k t} \right\|_2^2 \right\} \\ s.t. \sum_{k=1}^{K} v_k = f \end{cases}$$

(1)

式中：$\{v_k\}$ 表示模态分量；$\{\omega_k\}$ 表示中心频率；$\delta(t)$ 为脉冲函数。

以上约束问题得到的 Lagrange 表达式展开得到[32]：

$$L(\{v_k\},\{\omega_k\},\lambda) = \alpha \sum_{k=1}^{K} \left\| \partial_t \left[ (\delta(t) + j/\pi t) * v_k(t) \right] e^{-j\omega_k t} \right\|_2^2$$
$$+ \left\| f(t) - \sum_{k=1}^{k} v_k(t) \right\|_2^2 + \left\langle \lambda(t), f(t) - \sum_{k=1}^{k} v_k(t) \right\rangle$$

(2)

利用交替方向乘子搜寻增广 Lagrange 函数的鞍点。交替寻优迭代后的 $v_k$，$\omega_k$ 和 $\lambda$ 的表达式如下：

$$\hat{u}_k^{n+1}(\omega) = \frac{\hat{f}(\omega) - \sum_{i=1,i\neq k} \hat{u}(\omega) + \frac{\hat{\lambda}(\omega)}{2}}{1 + 2\alpha(\omega - \omega_k)^2}$$

(3)

$$\omega_k^{n+1} = \frac{\int_0^\infty \omega |\hat{v}_k(\omega)|^2 d\omega}{\int_0^\infty |\hat{v}_k(\omega)|^2 d\omega}$$

(4)

$$\hat{\lambda}^{n+1}(\omega) = \hat{\lambda}^n(\omega) + \gamma \left( \hat{f}(\omega) - \sum_k \hat{v}_k^{n+1}(\omega) \right)$$

(5)

其停止迭代准则为

$$\sum_k \left\| \hat{u}_k^{n+1} - \hat{u}_k^n \right\|_2^2 / \left\| \hat{u}_k^n \right\|_2^2 < \varepsilon$$

(6)

式中：$\varepsilon$ 为收敛阈值[33]。

## 1.2 模态分量高低频分界点确定

互信息是源于信息论中熵的概念，用来反映两个随机变量间的统计相关性[34-37]。在信息论中，信息并不是单独存在的，两个变量互信息越多，则相关性越大；反之，则越少。对于每个本征模态分量可计算其信息熵为

$$H(imf_n) = -\sum_i p(imf_n) \lg(p(imf_n))$$

(7)

式中：$p(imf_n)$ 为第 $n$ 个模态分量的发生概率。两个相邻的模态分量之间的互信息熵为：

$$I(imf_n, imf_{n+1}) = H(imf_n) + H(imf_{n+1}) - H(imf_n, imf_{n+1})$$

(8)

根据信息论可知，两个随机变量相互独立，其信息熵应等于零[38-40]。由此可知，相邻模态分量的信息熵从低频到高频排列存

在极小值点，此点即为高低频分界点。

## 2 极限学习机

极限学习机具有参数设置少、计算速度快、结构简单等优势[41]。其网络模型结构下所示：

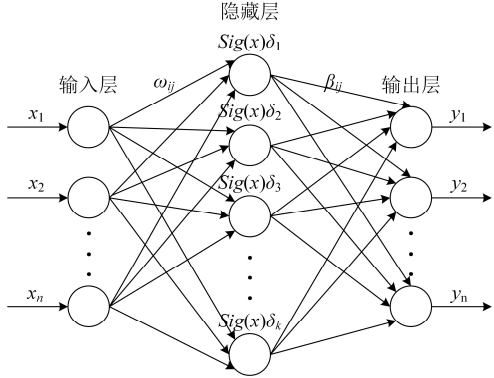

图 1 极限学习机网络模型

Fig.1 Extreme Learning Machine Network Model

设训练集样本数据为（$x_i, y_i$），其中 $x_i$ = [$x_{i1}, x_{i2}, \cdots, x_{in}$]为训练集 $n$ 维输入数据。$t_i$ = [$t_{i1}, t_{i2}, \cdots, t_{im}$]为训练集的 $m$ 维理想预测值。结合输入输出层权值、隐藏层阈值及激活函数可得到极限学习机网络模型结构数学表达式：

$$y_j = \sum_{i=1}^{K} \beta_i Sig_i\left(\omega_i \cdot x_j + \delta_i\right), j=1,2,\cdots,K \quad (9)$$

式中：$\omega_i$ 为输入层至隐藏层权重；$\beta_i$ 为隐藏层至输出层权重；$\delta_i$ 为隐藏层阈值；$Sig_i$ 为激活函数；$y_j$ 为网络模型结构的实际输出。

令上述极限学习机输出能零误差无限接进任意 $N$ 个训练样本[42]，即

$$\sum_{i=1}^{N} \|y_i - t_i\| = 0 \quad (10)$$

将上式代入式（1）

$$t_j = \sum_{i=1}^{K} \beta_i Sig_i\left(\omega_i \cdot x_j + \delta_i\right), j=1,2,\cdots,N$$

$$(11)$$

再将上式化简可得到

$$H \cdot \beta = T \quad (12)$$

式中：$H$ 为隐藏层输出矩阵；$T$ 为理想的输出向量。

最后,求解最小二乘解可得[43]：

$$\hat{\beta} = H^+ \cdot T \quad (13)$$

式中：$H^+$ 为 $H$ 矩阵的增广逆矩阵。

## 3 VMD-IPSO-ELM 预测模型

### 3.1 粒子群优化算法

PSO 算法由 J. Kennedy 和 R. C. Eberhart 在 1995 年提出，它是一种受到人类和动物群体行为启发的仿生算法[44]。在 PSO 中，每个解决方案都被视为一个"粒子"，而每个粒子的位置代表了一个潜在的解。这些解的质量由一个适应度函数来评估。粒子们依靠迭代过程中的信息更新自己的搜索方向和距离，以此找到问题的最优解。粒子的移动速度决定了其搜索路径的方向和长度。每个粒子都会记录自己找到的最佳位置 $P_{best}$，种群中所有粒子的最优解称为种群全局最优解 $G_{best}$，在每一次迭代中，每个粒子都会根据自己的个体最优和群体的全局最优信息来调整自己的速度，从而更新自己的位置。粒子位置的更新方法如下式所示[45-47]。

$$x_i^{t+1} = x_i^t + v_i^{t+1} \quad (8)$$

$$v_i^{t+1} = \omega v_i^t + c_1 r_1\left(P_{best,i}^t - x_i^t\right) + c_2 r_2\left(G_{best,i}^t - x_i^t\right) \quad (9)$$

式中：$x_i^{t+1}$ 为第 $t+1$ 次迭代时粒子 $i$ 的位置；$v_i^{t+1}$ 为第 $t+1$ 次迭代时粒子 $i$ 的速度；$\omega$ 为惯性权重因子；$c_1$ 和 $c_2$ 分别为个体和种群学习因子；$r_1$ 和 $r_2$ 分别为[0,1]内的随机数；$P_{best}^t{}_{,i}$ 和 $G_{best}^t{}_{,i}$ 分别为第 $t$ 次迭代时的个体最优解和种群全局最优解。

### 3.2 改进 PSO 算法

研究指出，种群初始化是智能优化算法中的关键步骤，初始化的质量直接影响到算法的收敛速度和鲁棒性[48]。在传统的 PSO 中，常常采用随机初始化种群的方法。然而，当种群规模较小或搜索空间维度较大时，随机初始化的分布均匀性往往较差，无法有效覆盖整个解空间。

混沌理论是研究非线性动态系统的一个分支，其特点包括遍历性、随机性以及对初值的高度敏感性[49]。采用混沌序列进行种群初始化可以实现更均匀的分布和较高的多样性，有助于算法跳出局部最优，提升收敛速度。传统的 Tent 混沌映射由于其分段性质，虽然能保证点的均匀分布，但同时也使得相邻点间具有较强的相关性，这可能导

致算法在小循环或不动点上陷入问题，尤其是当最优解只能在边缘值时[50]。针对 Tent 混沌映射存在的问题，学者们已经提出了多种有效的改进措施。其中一种典型的改进是在生成 Tent 混沌序列的过程中引入一个随机序列作为参数变量。这种方法通过随机信号的干扰来打破相邻点之间的相关性，并避免算法陷入短周期或不动点的重复迭代[51-52]。基于这些考虑，通过在传统 Tent 映射的公式中加入符合 beta 分布的随机数，可以获得一种结合了随机信号的改进混沌映射形式，旨在增强算法的全局搜索能力并提高其解的质量。

$$y_i^{j+1}=\begin{cases}\eta\times y_i^j+\gamma*brd(m,n) & y_i^j<0.5\\ \eta\times(1-y_i^j)+\gamma*brd(m,n) & y_i^j\geq 0.5\end{cases} \quad(19)$$

式中，$\eta$ 表示混沌系数，其值越大，混沌性能越优，本文取 $\eta$ 为 2；$brd$ 为 MATLAB 中的随机数生成器，可以生成符合 beta 分布的随机数，$\gamma$ 为收缩因子，用于对初始 Tent 种群进行扰动，$\gamma$、$m$ 和 $n$ 分别取 0.1、3 和 4。最后对混沌序列进行逆映射，即可得到初始种群个体的位置变量为

$$x_i^j = lb_j + y_i^j \times (ub_j - lb_j) \quad(20)$$

## 4 VMD-IPSO-ELM 组合模型预测流程

为提升负荷预测的精确度，本研究提出了一种结合 VMD 和 IPSO-ELM 的组合预测模型。该模型首先使用 VMD 算法对原始数据进行分解，提取不同频率的模态分量。通过互信息熵的方法评估各分量之间的相关性，并将这些模态分量重组为高频和低频序列[53-55]。接下来，利用 IPSO-ELM 模型对这两种序列分别进行负荷预测。最终，通过将预测的高频和低频负荷数据重构组合，形成预测结果[56-58]。IPSO-ELM 的具体流程如下：

步骤 1：将 ELM 权重、阈值作为 PSO 种群，并采用改进 Tent 混沌映射初始化 PSO 种群。

步骤 2：设定 PSO 相关参数，如粒子的初始速度、位置、训练次数和规模等。

步骤 3：进入迭代循环，并设置每一代的学习因子跟随同代的惯性权重变化。

步骤 4：确定粒子的适应度函数。本文将预测模型的性能指标值作为粒子的适应度函数，寻找最优模型参数。

步骤 5：计算粒子的适应度值，更新最优的适应度值。

步骤 6：终止准则判断，若达到最大迭代则输出最优预测精度下的 ELM 相关参数。若未达到要求，则继续迭代。

## 5 仿真分析

本研究以 2019 年 10 月份吉林省某风电场的实测数据为基础，数据采样间隔为 15 分钟，总共包含 1096 个数据点。研究中将数据集的前 75%作为训练样本集进行模型训练，剩余的 25%用作测试样本集以评估模型性能[59-61]。本文首先对基于 IPSO-ELM 预测模型进行了验证，并将其与传统粒子群优化算法以及未经任何参数优化的极限学习机进行了对比分析。各算法参数设置见表 1。

表 1 算法参数
Tab.1 Algorithm parameters

| 算法 | 参数 1 | 参数 2 | 参数 3 | 参数 4 |
|---|---|---|---|---|
| ELM | 输入层神经元 | 隐藏层数 | 激活函数 | 输出层神经元 |
|  | $N_{in}=7$ | $M=40$ | $Sigmoid$ | $N_{out}=1$ |
| PSO | 学习因子 1 | 学习因子 2 | 惯性权重 | — |
|  | $c_1=1.5$ | $c_2=1.5$ | $\omega=0.8$ |  |

为避免随机性对预测结果产生影响，取 30 次实验的预测结果的平均绝对百分误差最大、最小及平均值作为评价指标，衡量不同模型的预测精度[62-64]。图 2 为不同预测模型下的的预测结果对比，表 2 为不同算法的平均绝对百分误差对。

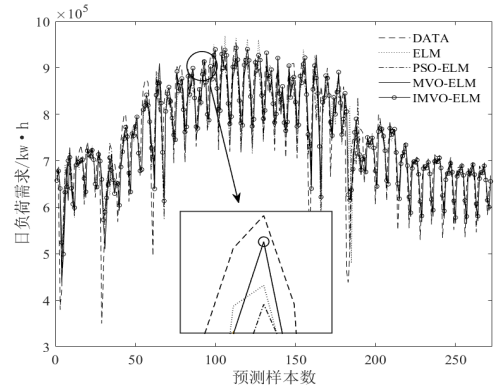

图 2 预测结果对比
Fig.2 Comparison of forecast results

表 2 预测误差对比
Tab.2 Prediction Error Comparison

| 优化算法 | MAPE$_{max}$ | MAPE$_{min}$ | MAPE$_{mean}$ |
| --- | --- | --- | --- |
| ELM | 7.41 | 5.06 | 6.73 |
| PSO-ELM | 5.78 | 5.27 | 5.44 |
| IPSO-ELM | 4.14 | 2.83 | 3.29 |

在上述预测结果中，未经参数优化的 ELM 展示了最高的平均 MAPE 值，显示出预测精度的不稳定性。虽然其最低 MAPE 值优于通过 PSO 优化得到的最低 MAPE 值，这主要得益于其随机初始化权值和阈值的特点[65-66]。在本研究提出的改进粒子群优化极限学习机（IPSO-ELM）组合预测模型中，平均 MAPE 值相比 ELM 和 PSO-ELM 模型分别降低了 1.93%和 2.15%，而预测精度则分别提高了 36.97%和 39.52%。同时，该模型在最大和最小 MAPE 值上也都表现出最低值，显示出比其他预测模型更优的鲁棒性和预测精度，从而确认了 IPSO-ELM 算法在预测效果上的优越性。

根据分析可验证 IPSO-ELM 预测模型的有效性。接下来则借助 VMD 算法对历史数据的预处理，进一步验证 VMD-IPSO-ELM 组合预测模型的优越性。原始负荷数据如图 3 所示

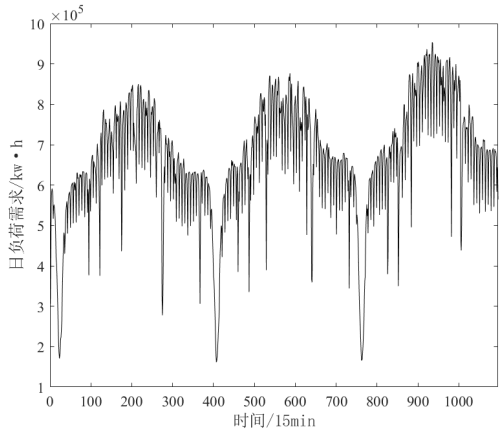

图 3 原始负荷数据
Fig.3 original load data

由图 3 可知，原始负荷需求数据整体具有明显的不稳定性，且极个别点幅值波动较大。若直接进行负荷预测，结果往往存在较大的预测误差。因此通过 VMD 算法，取分解层数 K 为 7，将原始数据分解为频率由低到高排列的模态分量，如图 4 所示。

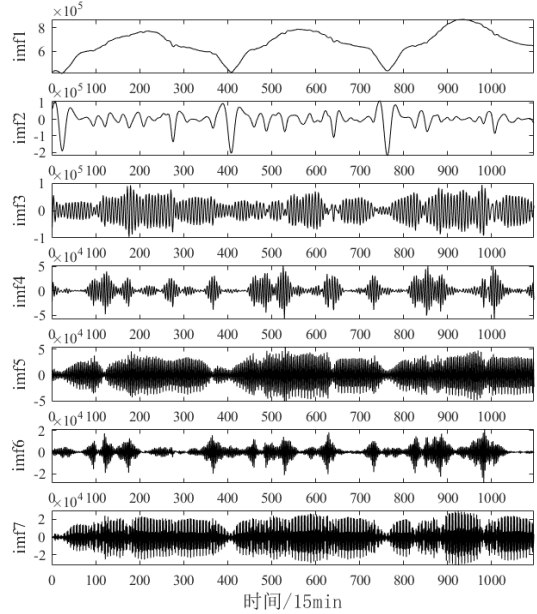

图 4 IMF 模态分量
Fig.4 IMF modal components

在 VMD 算法的分解下，各模态分量的中心频率均能得到有效分离。然后通过互信息熵值衡量不同模态分量间的相关性，取互信息熵出现的极小值点处作为模态分量的分界点，进而得到高频、低频序列[67-68]。最后分别将高频、低频序列作为 IPSO-ELM 预测模型输入进行负荷预测，并将两者预测结果重构得到最后预测数据[69]。所提组合预测结果如图 5 所示，预测误差如表 3 所示。

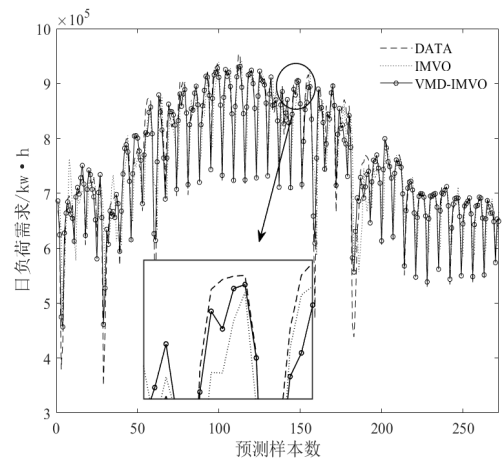

图 5 VMD-IPSO-ELM 组合预测结果
Fig.5 VMD-IPSO-ELM combined prediction results

表 3 组合预测误差
Tab.3 Combined forecast error

| 预测算法 | MAPE (%) | RMSE(kw·h) |
|---|---|---|
| IPSO | 3.46 | 5.91 |
| VMD-IPSO | 2.03 | 2.83 |

结合图 9 及表 3 可知，VMD-IPSO-ELM 组合预测模型相比单一预测模型 IPSO-ELM。其预测结果的平均绝对百分误差又进一步降低至 2.03%，预测精度在原有基础上提高 41.3%，均方根误差降低了 3.08 kW·h。这表明通过 VMD 算法将原始负荷划分为高、低频序列的数据预处理可以有效降低数据波动对预测结果的影响，提高预测精度[70]。此外，这一分析也进一步验证了所提组合预测模型的优越性。

## 6 结论

为提升电力负荷预测的准确性，本文从优化算法和数据预处理两个关键方面入手，研发了基于 VMD 和 IPSO-ELM 的先进组合预测模型。仿真结果表明，采用 IPSO-ELM 模型在寻优能力、收敛速度和预测精度等方面展现出卓越的性能。此外，通过变分模态分解对原始电力数据进行预处理，分离出的高频和低频序列经独立预测和后续重构，使得最终的预测精度得到了显著提高。

此外，本研究进一步强调了数据预处理在提高预测模型性能中的重要作用。VMD 算法不仅能有效分离高频和低频信息，而且通过这种分解增强了模型对数据内在动态的捕捉能力，从而在实际应用中更加精准地预测电力负荷波动。本研究证实了所提预测模型不仅具有理论上的创新性，显著提高了电力负荷预测的准确率和可靠性，同时在电力系统优化调度中也具有较高的应用价值[71-72]。此外，研究信息攻击情形下负荷预测[73-74] 和采用深度迁移学习[75-76]进行负荷预测，都是未来值得深入研究的方向。

**参考文献**

[1] 高翱, 王帅, 韩兴臣, 等. 基于 GRU 神经网络的 WGAN 短期负荷预测模型[J]. 电气工程学报，2022, 17(2): 168-175.

[2] 杨健, 孙涛, 陈小龙, 等. 基于 VMD-EWT-IASSP-EBILSTM 的短期电力负荷预测[J]. 科学技术与工程, 2023, 23(27): 11646-11654.

[3] Kou L, Wu J, Zhang F, et al. Image encryption for Offshore wind power based on 2D-LCLM and Zhou Yi Eight Trigrams[J]. International Journal of Bio-Inspired Computation, 2023, 22(1): 53-64.

[4] Li Y, Yang Z, Li G, et al. Optimal scheduling of an isolated microgrid with battery storage considering load and renewable generation uncertainties[J]. IEEE Transactions on Industrial Electronics, 2019, 66(2): 1565-1575.

[5] 曹敏, 李文云, 钱详华, 等. 基于分类识别深度置信网络的电力负荷预测算法[J]．电力需求侧管理，2020，22(02)：44-50．

[6] Zhang S, et al. A critical review of data-driven transient stability assessment of power systems: principles, prospects and challenges[J]. Energies, 2021, 14(21): 7238.

[7] 周磊, 竺筱晶. 基于 MA-CNN-LSTM 和自注意力机制的 单变量短期电力负荷预测 [J]. 科学技术与工程，2024, 24(22).

[8] Shi Z, Li Y, Yu T. Short-term load forecasting based on LS-SVM optimized by bacterial colony chemotaxis algorithm[C]//2009 International Conference on Information and Multimedia Technology. IEEE, 2009: 306-309.

[9] Bentouati B, Chettih S, Jangir P, et al. A solution to the optimal power flow using multi-verse optimizer[J]. Journal of Electrical Systems, 2016, 12(4).

[10] Li Y, Gu X P. Application of online SVR in very short-term load forecasting[J]. International review of electrical engineering, 2013, 8(1): 277-282.Mirjalili

S, Mirjalili S M, Hatamlou A. Multi-Verse Optimizer: a nature-inspired algorithm for global optimization[J]. Neural Computing & Applications, 2016, 27(2):495-513.

[11]黄裕春, 贾巍, 雷才嘉, 等. 基于混沌多目标蚁狮优化算法和核极限学习机的冲击性负荷预测模型[J]. 现代电力, 2023, 40(6): 1043-1051.

[12]黄志祥, 周莉. 基于 VMD-LSTM 的短期电力负荷预测研究[J]. 洛阳理工学院学报 (自然科学版), 2022, 32(3): 76-80.

[13]栗然，孙帆，丁星，等. 考虑多能时空耦合的用户级综合能源系统超短期负荷预测方法[J]. 电网技术，2020，44(11)：4121-4134.

[14]LAI Wenhao, ZHOU Mengran, HU Feng, et al. A new DBSCAN parameters determination method based on improved MVO[J]. IEEE Access, 2019, 7: 104085–104095.

[15]刘岩，彭鑫霞，郑思达. 基于改进 LS-SVM 的短期电力负荷预测方法研究[J]. 电测与仪表，2021，58(05)：176-181.

[16]Zhang Y, Li T, Na G, et al. Optimized extreme learning machine for power system transient stability prediction using synchrophasors[J]. Mathematical Problems in Engineering, 2015, 2015(1): 529724

[17]Dragomiretskiy K, Zosso D. Variational Mode Decomposition[J]. IEEE Transactions on Signal Processing, 2014, 62(3): 531-544.

[18]DU Jiani, LIU Zhitao, WANG Youyi. State of charge estimation forLi-ion battery based on model from extreme learning machine[J].Control Engineering Practice, 2014, 26: 11

[19]刘栋, 魏霞, 王维庆,等. 基于 SSA-ELM 的短期风电功率预测[J]. 智慧电力, 2021, 49(6):8.

[20]Li Y, Li K, Yang Z, et al. Stochastic optimal scheduling of demand response-enabled microgrids with renewable generations: An analytical-heuristic approach[J]. Journal of Cleaner Production, 2022, 330: 129840.

[21]Yahui Li, Yang Li, Guoqing Li. Optimal power flow for AC/DC system based on cooperative multi-objective particle swarm optimization[J]. Automation of Electric Power Systems, 2019, 43 (4): 94-100.

[22]Wang Y, Wang Z, Wang G G. Hierarchical learning particle swarm optimization using fuzzy logic[J]. Expert Systems with Applications, 2023, 232: 120759.

[23]Wang L, Yang Y, Xu L, et al. A particle swarm optimization-based deep clustering algorithm for power load curve analysis[J]. Swarm and Evolutionary Computation, 2024, 89: 101650.

[24]王金玉, 胡喜乐, 闫冠宇. 基于 VMD 的 CNN-BiLSTM-Att 的短期负荷预测[J]. 吉林大学学报 (信息科学版), 2023, 41(6): 1007-1014.

[25]Tharwat A, Hassanien A E. Quantum-behaved particle swarm optimization for parameter optimization of support vector machine[J]. Journal of Classification, 2019, 36(3): 576-598.

[26]Li Y, Feng B, Li G, et al. Optimal distributed generation planning in active distribution networks considering integration of energy storage[J]. Applied energy, 2018, 210: 1073-1081.

[27]封磊, 蔡创, 齐春, 等. PSO 和 GA 的对比及其混合算法的研究进展[J]. 控制工程, 2005 (S2): 93-96.

[28]陈建华, 李先允, 邓东华, 等. 粒子群优化算法在电力系统中的应用综述[J]. 电力系统保护与控制, 2007, 35(23): 77-84.

[29]兰华, 艾涛, 等. 经验模态分解在单相自适应重合闸中的应用[J]. 电力系统保护与控制, 2010, 38(12): 35-39.

[30]Cui J, Jin Y, Yu R, et al. A robust approach for the decomposition of

high-energy-consuming industrial loads with deep learning[J]. Journal of Cleaner Production, 2022, 349: 131208.

[31]江星星, 宋秋昱, 杜贵府, 等. 变分模式分解方法研究与应用综述[J]. 仪器仪表学报, 2023 (1): 55-73.

[32]赵凤展, 郝帅, 张宇, 等. 基于变分模态分解-BA-LSSVM 算法的配电网短期负荷预测[J]. 农业工程学报, 2019, 35(14): 190-197.

[33]张莲, 李涛, 宫宇. 考虑变分模态分解残差量的电力负荷预测研究[J]. 重庆理工大学学报（自然科学）, 2022, 36(1): 165-170.

[34]Acharya Srinivasa et al. A multi-objective multi-verse optimization algorithm for dynamic load dispatch problems[J]. Knowledge-Based Systems, 2021, 231

[35]Poles S, Fu Y, Rigoni E. The Effect of Initial Population Sampling on the Convergence of Multi-Objective Genetic Algorithms[J]. LECTURE NOTES IN ECONOMICS AND MATHEMATICAL SYSTEMS, 2009.

[36]刘建新, 李朝伟, 张楷生. 一种新的改进型 Tent 混沌映射及其性能分析[J]. 科学技术与工程, 2013(8):5.

[37]张娜,赵泽丹,包晓安,钱俊彦,吴彪.基于改进的 Tent 混沌万有引力搜索算法[J]. 控制与决策,2020,35(04):893-900.

[38]段雪滢, 李小腾, 陈文洁. 基于改进粒子群优化算法的 VMD-GRU 短期电力负荷预测[J]. 电工电能新技术，2022, 41(5): 8-17.

[39]郭傅傲,刘大明,张振中,唐飞.基于特征相关分析修正的 GPSO-LSTM 短期负荷预测[J].电测与仪表,2021,58(6):39-48.

[40]臧淑娟．用电负荷大数据分析预测系统的设计与实现[D]．山东大学，2018．

[41]Tang, Z., et al. Data driven based dynamic correction prediction model for NOx emission of coal fired boiler[J]. Proceedings of the CSEE, 2022, 42, 5182-5193.

[42]Li Y, Yang Z. Application of EOS-ELM with binary Jaya-based feature selection to real-time transient stability assessment using PMU data[J]. IEEE Access, 2017, 5: 23092-23101.

[43]Tang Z, Wang S, Chai X, et al. Auto-encoder-extreme learning machine model for boiler NOx emission concentration prediction[J]. Energy, 2022, 256: 124552.

[44]王煜尘, 窦银科, 孟润泉. 基于模糊 C 均值聚类-变分模态分解和群智能优化的多核神经网络短期负荷预测模型[J]. 高电压技术, 2022, 48(4): 1308-1319.

[45]冯茜, 李擎, 全威, 等. 多目标粒子群优化算法研究综述[J]. 工程科学学报，2021, 43(6): 745-753．

[46]朱丽莉, 杨志鹏, 袁华. 粒子群优化算法分析及研究进展[J]. 计算机工程与应用, 2007, 43(5): 24-27．

[47]曾裕钦, 蔡华洋, 周茹平, 等. 基于混合离散粒子群优化的控制模式分配算法[J]. 电子学报, 2024: 1-14．

[48]Li Y, Li Y, Li G, et al. Two-stage multi-objective OPF for AC/DC grids with VSC-HVDC: Incorporating decisions analysis into optimization process[J]. Energy, 2018, 147: 286-296.

[49]Shi Z, Yu T, Zhao Q, et al. Comparison of algorithms for an electronic nose in identifying liquors[J]. Journal of Bionic Engineering, 2008, 5(3): 253-257.

[50]顾雪平, 等. 基于局部学习机和细菌群体趋药性算法的电力系统暂态稳定评估[J]. 电工技术学报，2013，28(10): 271-279．

[51]朱敏, 王石记, 杨春玲. 改进 Tent 混沌序列的数字电路 BIST 技术[J]. 哈尔滨工业大学学报, 2010, 42(4): 607-618．

[52]Li Y, Feng B, Wang B, et al. Joint planning of distributed generations and energy storage in active distribution networks: A Bi-Level programming approach[J]. Energy, 2022, 245: 123226

[53] 杨晶显，张帅，刘继春，等．基于 VMD 和双重注意力机制 LSTM 的短期光伏功率预测[J]．电力系统自动化，2021，45(03)：174-182．

[54] MAGNUS D，ADAM B，OLIVER S K，Gorm B A．Improving short-term heat load forecasts with calendar and holiday data[J]．Energies，2018，11(7)：1678．

[55] AMARASINGHE K，MARINO D L，MANIC M．Deep neural networks for energy load forecasting[C]．2017 IEEE 26th International Symposium on Industrial Electronics（ISIE）．IEEE，2017：1483-1488．

[56] 段明明，杨捷，李沛霖．基于小波和径向基函数神经网络的电力负荷预测研究[J]．云南大学学报(自然科学版)，2020，42(S2)：18-25．

[57] LAI C S，YANG Y，PAN K，et al．Multi-view neural network ensemble for short and mid-term load forecasting[J]．IEEE Transactions on Power Systems，2021，36(4)：2992-3003．

[58] APRILLIA H，YANG H T，HUANG C M．Statistical load forecasting using optimal quantile regression random forest and risk assessment index[J]．IEEE Transactions on Smart Grid，2020，Early Access(99)．

[59] 刘建华，李锦程，杨龙月，等．基于 EMD-SLSTM 的家庭短期负荷预测[J]．电力系统保护与控制，2019，047(006)：40-47．

[60] Fang Z, Zhao D, Chen C, et al. Nonintrusive appliance identification with appliance-specific networks[J]. IEEE Transactions on Industry Applications, 2020, 56(4): 3443-3452．

[61] 邓带雨，李坚，张真源，等．基于 EEMD-GRU-MLR 的短期电力负荷预测[J]．电网技术，2020，44(02)：593-602．

[62] Deng Daiyu，Li Jian，Zhang Zhenyuan，et al．Short-term electric load forecasting based on EEMD-GRU-MLR[J]．Power System Technology，2020，44(02)：593-602．

[63] 高金兰，王天．基于 VMD-IWOA-LSSVM 的短期负荷预测[J]．吉林大学学报 (信息科学版), 2021, 39(4): 430-438．

[64] HE K，ZHANG X，REN S，et al．Deep residual learning for image recognition[C]// IEEE Conference on Computer Vision & Pattern Recognition．IEEE Computer Society，2016：770-778．

[65] HE K，ZHANG X，REN S，et al. Identity mappings in deep residual networks[C]// European Conference on Computer Vision．Springer International Publishing，2016：630-645．

[66] KO M，LEE K，KIM J K，et al．Deep concatenated residual network with bidirectional LSTM for one-hour-ahead wind power forecasting[J]．IEEE Transactions on Sustainable Energy，2021，12(2)：1321-1335．

[67] 伍骏杰, 张倩, 陈凡, 等. 计及误差修正的 VMD-LSTM 短期负荷预测[J]. 科学技术与工程, 2022, 22(12): 4828-4834．

[68] 陆磊，张铭飞，朱浩钰．基于 VMD-IPSO-LSSVM 的短期电力负荷预测研究[J]．电工技术，2023 (24): 175-178．

[69] 穆晨宇, 薛文斌, 穆羡瑛, 等. 基于 VMD-LSTM-Attention 模型的短期负荷预测研究[J]. 现代电子技术, 2023, 46(17): 174-178．

[70] 赵星宇，吴泉军，朱威．基于 CEEMDAN 和 TCN-LSTM 模型的短期电力负荷预测[J]. 科学技术与工程, 2023, 23(4): 1557-1564.

[71] Li Y, Han M, Yang Z, et al. Coordinating flexible demand response and renewable uncertainties for scheduling of community integrated energy systems with an electric vehicle charging station: A bi-level approach[J]. IEEE Transactions on

Sustainable Energy, 2021, 12(4): 2321-2331.

[72] Qu Z, Dong Y, Mugemanyi S, et al. Dynamic exploitation Gaussian bare‐bones bat algorithm for optimal reactive power dispatch to improve the safety and stability of power system[J]. IET Renewable Power Generation, 2022, 16(7): 1401-1424.

[73] Qu Z, Zhang Y, Qu N, et al. Method for quantitative estimation of the risk propagation threshold in electric power CPS based on seepage probability[J]. IEEE Access, 2018, 6: 68813-68823.

[74] Wang L, et al. Method for extracting patterns of coordinated network attacks on electric power CPS based on temporal–topological correlation[J]. IEEE Access, 2020, 8: 57260-57272.

[75] Li Y, Zhang S, Li Y, et al. PMU measurements-based short-term voltage stability assessment of power systems via deep transfer learning[J]. IEEE Transactions on Instrumentation and Measurement, 2023, 72: 2526111.

[76] Santos M L, García S D, García-Santiago X, et al. Deep learning and transfer learning techniques applied to short-term load forecasting of data-poor buildings in local energy communities[J]. Energy and Buildings, 2023, 292: 113164.